\documentclass[a4paper,fleqn]{cas-dc}

\usepackage{lipsum}

\usepackage[numbers]{natbib}
\usepackage{float} 

\xspaceaddexceptions{]}
\def\tsc#1{\csdef{#1}{\textsc{\lowercase{#1}}\xspace}}
\tsc{WGM}
\tsc{QE}
\tsc{EP}
\tsc{PMS}
\tsc{BEC}
\tsc{DE}

\begin{document}
\let\WriteBookmarks\relax
\def\floatpagepagefraction{1}
\def\textpagefraction{.001}
\shorttitle{Locally interacting SOM}

\shortauthors{Siddiqui \& Georgiadis}

\title [mode = title]{Global Collaboration through Local Interaction in Competitive Learning}

\author[add1]{Abbas Siddiqui}
\ead{abbas.siddiqui@frs.ethz.ch}
\author[add1]{Dionysios Georgiadis}
\ead{dionysios.georgiadis@frs.ethz.ch}

\address[add1]{Future Resilient Systems, ETH Zurich}

\begin{abstract}
Feature maps, that preserve the global topology of arbitrary datasets, can be formed by self-organizing competing agents. So far, it has been presumed that global interaction of agents is necessary for this process. We establish that this is not the case, and that global topology can be uncovered through strictly local interactions. Enforcing uniformity of map quality across all agents, results in an algorithm that is able to consistently uncover the global topology of diversely challenging datasets.The applicability and scalability of this approach is further tested on a large point cloud dataset, revealing a linear relation between map training time and size. The presented work not only reduces algorithmic complexity but also constitutes first step towards a distributed self organizing map.
\end{abstract}

\maketitle

\section{Introduction}\label{:introduction}

The Self Organizing Map (SOM) is a competitive, unsupervised learning algorithm capable of creating a low dimensional and discrete representation of high dimensional data. 

Since its initial conception, SOM has found broad application in data analytics, mainly for data clustering, function approximation, and dimensionality reduction (see \cite{kohonen2013_essentials, kohonen_book} for examples of applications).

A SOM consists of a population of adaptive, interacting agents dubbed \textit{units}. Each unit is represented in the sample space by vector (called weight) and it influences set of other units (\textit{neighbors}).
For each sample, the unit with the most similar weight is found (called the \textit{best matching unit - BMU}), and its similarity to the sample is increased by altering its weight. Subsequently, the neighbors of the BMU are also influenced by increasing their similarity to the sample - albeit to a lesser extend.

Given enough data, the units' weight may converge to a low dimensional discrete representation of the data - called a feature map. 
Additionally, the units' weight will be placed meaningfully: neighboring units should contain similar features, since neighborhoods move en masse, a property known as topological preservation.
It is possible however for this process to go awry; for example, limiting the influence of a unit over its neighbors may compromise the topological preservation\cite{kohonen2013_essentials,keith1999_empirical}. For reference, we dub this phenomenon \textit{topological deformation}.

Topological deformation in SOM is typically dealt with by using larger neighborhood size - an empirically established treatment as stated in \cite{kohonen2013_essentials}. 
However, larger neighborhoods also increase the algorithm's \textit{computational complexity} in proportion to their size, thus requiring significant computational resources and restricting the scalability.
At the best of the authors' knowledge, alternative methods for resolving topological frustration have not been investigated and the large neighborhoods are always used  - in spite of their cost.
Furthermore, so far no study has systematically focused on the phenomenology of the SOM in the limit of very small neighborhoods. 
With little to no understanding of the specific problems that stem from very small neighborhoods, it is arguably impossible to investigate solutions. 

To address these gaps, we i) investigate the pathology of SOM in the limit of small neighborhoods, and ii) propose an alternative treatment, with drastically smaller computational complexity than larger neighborhoods. The treatment utilizes localized feedback loop to enforce uniformity in the map errors. The efficacy of the approach is tested empirically on synthetic data. Finally, the applicability and the scalability are investigated empirically in a SOM application: point cloud estimation\cite{pointcloud_1,pointcloud_2,pointcloud_3}.

 With the current work we illustrate a promising paradigm for SOM: harnessing the dynamics of locally interacting adaptive systems in order to replace large neighborhoods by computationally efficient alternatives.

Doing so will not only reduce computational complexity, but will also result in looser coupling, which is a first step towards a fully distributed version of SOM.
Such improvements enable new applications where performance is paramount such as on-line learning over big data streams or data discovery using very large maps\cite{massivemap}.

\section{Background}\label{:GeneralImp}
Training an SOM consists of two separate processes: finding the best matching unit and adjusting the map. 
The map units are arranged as a lattice graph $G$, which is a square lattice in the presented work.

\textbf{Best Matching Unit (BMU):} Given the $i$th training sample $s_i$, the BMU is found by comparing the $s_i$ distance from all the units position $w_j(i)$ , as shown in eq. \eqref{bmueq}. Usually, the Euclidean distance is used to find the BMU.	
\begin{align}
	b_i = argmin_{ 1 \leq j \leq n} \| w_j(i) - s_{i} \| \label{bmueq} 	
\end{align}

Where $n$ is number of units in the map, and $b_i$ is the index of the best matching unit of the $i$th sample. Whereas, $w_j(i)$  is a vector denoting the $j$th unit's position in the sample space on the $s_i$.

In our analysis, we will also make use of the second BMU, which is denoted by $\hat{b}_{i}$, and found by solving:
\begin{align} \label{secondbmu}	
	\hat{b}_{i} = {\arg \min}_{j \in \{ 1,..., n \} \setminus \{b_{i} \} } \|w_j(i) - s_{i}\|
\end{align}

\textbf{Map Adjustment:} The heart of the SOM is the adjustment of the BMU and its neighbors according to a training sample. The strength of the adjustment decays with respect to the time (i.e., in terms of number of samples processed) and the distance (between a unit and the BMU). This distance is typically measured by the number of hops needed to move from a unit $j$ to the BMU over the graph $G$. We refer to this distance as $D_{j,b_i}$. The adjustment rate of unit $j$ at sample $i$ is  $r_{{j}}(i)$, and can be given by any function that is strictly decreasing with respect to both $i$ and $D_{j,b_i}$. 

The adjustment of unit $j$ due to sample $i$ is given by:	
\begin{align}	
	\Delta w_j(i) = r_{{j}} \big(i \big) \big( s_i-w_j(i) \big) \label{pf2}	
\end{align}

\textbf{Quality Metrics:} 
Map coverage is quantified by the \textit{quantization error}\cite{KohonenQError} metric.
The quantization error of the $i$th sample is given by:
\begin{equation}\label{qerror}
	q(i) =  ||w_{b_i}(i) - s_{i}||
\end{equation}

Topological deformation is quantified using the \textit{alfa error}\cite{alfaError} metric. The alfa error is defined for square lattice SOMs, and relies on the concept of diagonal neighbors: the units in the Moore neighborhood but not in Von Neumann neighborhood of unit $j$.
The alfa error of the $i$th sample is calculated by comparing the positions of the two first BMUs $(b_i, \hat b_i)$ as defined below:
\begin{align} \label{aerror}	   
	\alpha(i) &=
	\begin{cases}
		0, & \text{ if $\hat{b}_{i}$ adjacent to $b_i$}  \\
		p, & \text{ if $\hat{b}_{i}$ diagonal neighbor of $b_i$}  \\
		1, & \text{ otherwise}  \\
	\end{cases} 
\end{align}
Where $p \in [0,1]$ is a user defined parameter, quantifying the alfa error in the case of the two BMUs being diagonal neighbors. 
For the current study the $p$ is set to $0.5$. 

\textbf{Local Map Quality:} For our analysis, it is needed to quantify the quality of the SOM on two separate scales: local, and global.
To measure the map quality on a local scale, we take the running means of the error values for each unit $j$:
\begin{subequations}
	\begin{align} \label{raerror}	   
		\bar q_j(i) &=
		\begin{cases}
			\eta \bar q_j(i-1) + q(i), & \text{ if $b_i=j$}  \\
			\bar q_j(i-1) , & \text{ otherwise}  \\
		\end{cases}  \\
		\notag \\
		\bar \alpha_j(i) &=
		\begin{cases}
			\eta \bar \alpha_j(i-1) + \alpha(i), & \text{ if $b_i=j$}  \\
			\bar \alpha_j(i-1) , & \text{ otherwise}  \\
		\end{cases}  \label{rqerror}	   
	\end{align}
\end{subequations}
Where $\bar q_j(i) , \bar \alpha_j(i)$ are the running means at sample $i$.
We initialize with $\bar q_j(0) =  \bar \alpha_j(0) = 0, ~ \forall j $.
The user defined parameter $\eta \in (1,0]$ controls the temporal decay of the running means, and it is set to $0.75$ throughout this work.	

\textbf{Global Map Quality}
To measure global map quality, we average over the unit errors given in \eqref{aerror} \eqref{qerror}.
For practical reasons, we measure training time not in number of samples but in \textit{iterations}, where one iteration is defined as $10 \cdot n$ samples. 
The linear map size dependence in the definition of one iteration allows us to compare the temporal evolution of the error metrics between maps of different sizes.
Specifically, for the $t$th iteration we have:
\begin{subequations}
	\begin{align}		
		Q_t = \frac{1}{n} \left. \sum_{j=1}^n \bar q_j (\tau) \right|_{\tau=n10t} \label{qmap}\\
		A_t = \frac{1}{n} \left. \sum_{j=1}^n \bar \alpha_j (\tau) \right|_{\tau=n10t}\label{alfamap}	\\ 
		A_{t} \text{ is the average of all $\bar{\alpha}_j(i)$ at iteration $t$} \notag \\		
		Q_{t} \text{ is the average of all $\bar{q}_j(i)$ at iteration $t$} \notag 	
	\end{align}

\end{subequations}

\section{Locally Interacting SOMs}

In classical SOM, the neighborhood attraction decreases with respect to time which effectively reduces the neighborhood size. The initial large (entire map - global) neighborhood size ensures global ordering whereas the eventual smaller (local) neighborhood size at the end of training phase ensures local order and stability of the map.

Reducing the neighborhood size to local for the entire training process will reduce computational complexity, but it is known to compromise global topological preservation as observed in \cite{kohonen2013_essentials, keith1999_empirical}.

\subsection{Constant Learning Rate(s)} \label{CLRSec}

To capture the topological deformation resulting from small neighborhoods, we experiment with locally interacting SOMs. Specifically, a neighborhood is constrained to a unit's immediate neighbors on graph $G$, see equation \eqref{LRAdj}, and learning rate is kept to user-defined constant $l_\zeta$. We refer to this SOM variant as Nearest Neighbors SOM (NNSOM).

	\begin{align}\label{LRAdj}
	r_{{j}} \big(i \big) &=
	\begin{cases}
	l_\zeta  e^{-D_{j,b_i}}  & \text{ if $D_{j,b_i} \leq 1$}  \\
	0 & \text{ otherwise}  
	\end{cases} 		
	\end{align}

We apply multiple individual NNSOMs over a dataset where points are uniformly scattered at random within a square. To illustrate the effect of the map size on the algorithm performance, we use two map sizes (400, 900).
The maps are trained for $3k$ iterations where each iteration consists of $n \cdot 10$ samples as defined in section \ref{:GeneralImp}.
The learning rate is constant in time, and takes 10 geometrically spaced values in $(0,1]$.
Constant learning rate is typically not used in SOM since there is no guarantee of  convergence \cite{LRMapConvergence} without temporally decaying learning rate. However, fixing the learning rate to a constant is particularly instructive when investigating the algorithm's potential in terms of topological preservation \cite{keith1999_empirical}, as it allows isolating the effects of other parameters on the algorithm.

Finally, two initializations are considered: \textit{curated symmetrical} (all unit weights placed at the origin) and \textit{random} (unit weights sprinkled uniformly at random over the dataset domain).
For the sake of brevity, we use the acronyms SIC and RIC to refer to Symmetric Initial Conditions and Random Initial Conditions respectively. Comparing the two cases allows us to quantify the sensitivity of the algorithm with respect to initial conditions.

 Each of the aforementioned configurations was run 20 times with random data points and in the case of RIC also with different initial conditions.

We assess the performance of NNSOM by observing the map alfa error and unit positions at iteration $3k$.
Remarkably, all observed maps fall into few classes of macroscopic behaviors, in spite of the large number of configurations tested in this experiment.
Furthermore, these behaviors are associated with different map alfa errors.

\begin{figure}
	\includegraphics[scale=0.40]{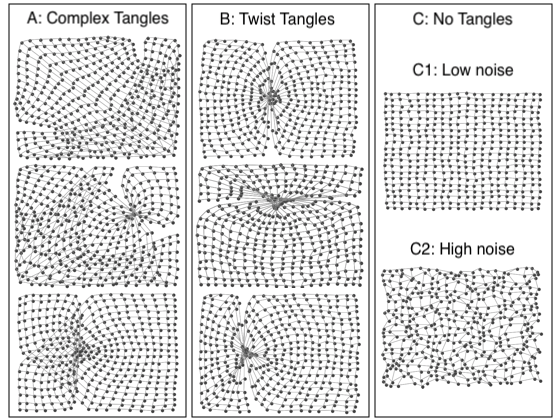}
	\caption{depicts different steady states of the NNSOM algorithm. The dots represent the map units, positioned in the sample space, after the maps reach a stationary state. Panels A and B depict maps with defective global topology unlike the maps in panel C. However, the map in C2 is of inferior local topological quality to C1. Note that the maps in panel B have twisted defects, while the defects in panel A are more complex.}\label{Regimes}
\end{figure}

\textbf{Robust Topological Defects:} We have identified an ad-hoc typology of macroscopic behaviors of NNSOM. Specifically, we assign each of the maps to one of three qualitatively distinct classes, and propose a name for each class -  as depicted in the three panels of figure \ref{Regimes}. Panels A and B depict the maps with incorrect global topology. Panel C shows the topologically preserved maps albeit with varying levels of noise.
Due to the tangled appearance of the maps in the panels A and B, these configurations are referred to as  \textit{tangles}.
The maps in panel A are twisted and folded onto themselves and therefore dubbed \textit{complex tangles}, while the maps in the panel B are labeled \textit{twist tangles}. Tangles were found to persist for thousand of iterations after their appearance and thus constitute \textit{robust topological defects}.
The map alfa error between tangled and untangled maps was found to differ up to a factor of 20.

\begin{figure}
	\centering
	\includegraphics[scale=0.40]{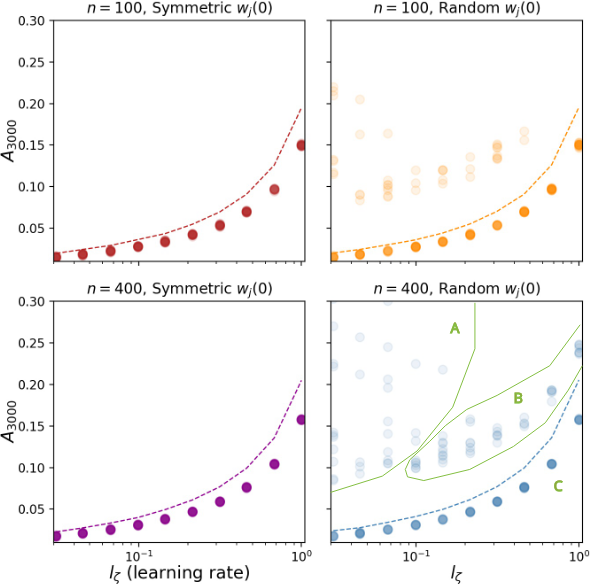}
	\caption{illustrates the map alfa error (eq.\eqref{alfamap}) at stationarity (iteration 3000).
		We show results for two map sizes, 10 learning rates, two initial conditions (random and curated). Each resulting combination is run 100 times over a square 2D dataset.
		For curated initial conditions, the values of map alfa error at iteration 3000 are tightly clustered around the same level.
		In contrast, for random initial conditions we observe number of points lying above the aforementioned locus. 
		We manually observed that these points correspond to topologically defective maps. 
		In the bottom right graph, we outline the identified map defects according to figure \ref{Regimes}.}\label{alfaCLR}
\end{figure}

The experiment reveals that both initial conditions and sizes have strong effect on the map alfa error as depicted in figure \ref{alfaCLR}.
The typology introduced in figure \ref{Regimes} allows a more intuitive interpretation of the algorithm's performance.

Specifically:

\begin{enumerate}
	\item
	For SIC, the values of $A_{3000}$ are tightly clustered together for each $l_\zeta$. However, $l_\zeta$ affects the level around which  $A_{3000}$ values are clustered. 

	Manually inspecting the unit positions in the sample space reveals that all maps with SIC are untangled, and that the variance of  $A_{3000}$ originates from varying levels of noise in the unit mesh (as shown in the right panel of figure \ref{Regimes}).
	\item
	Using RIC, we obtain a cloud of points in addition to the locus encountered for SIC. This cloud lies above the aforementioned locus. Manual inspection reveals that all the points in the cloud are tangled maps - revealing that the NNSOM is unable to resolve topological deformation of RIC consistently. 
	Furthermore, while in SIC lower $l_\zeta$ resulted in better map quality this is not the case in RIC: lowering $l_\zeta$ moves more points in the cloud of failed maps. 
	In fact, manual inspection revealed that fold tangles only appeared for sufficiently low $l_\zeta$ values. 
	To aid interpretation, the  $A_{3000}$ values that correspond to different map classes have been delineated in the bottom right panel of the figure \ref{alfaCLR} with green solid lines.
	\item
	For SIC, increasing the size of the map from 100 to 400 does not change the resulting locus of  $A_{3000}$ points, and thus does not have significant impact on the topological preservation.
	This is in contrast with RIC, where the larger map size not only increased the number of tangled maps, but also resulted in tangled maps even in the case of $l_\zeta=1$.
\end{enumerate}
In summary, while the NNSOM performs adequately well over a uniform square dataset for SIC, it may fail to resolve the topological frustration for RIC.
This shortcoming is exacerbated for increasing system size, and persists for any learning rate - and even after many iterations.
Applications of SOM generally involve more challenging problems than RIC over uniform square data, and therefore the NNSOM will not be applicable in such cases.
However, having identified the origin of NNSOMs shortcomings (formation of tangles), we will now consider a treatment.


\subsection{Feedback based Nearest Neighbors SOM} \label{fnnsomexp}

 Topological deformation results when a unit's neighbors are farther away than its non-neighbors. This issue can be observed in the center of symmetry of the twisted tangles maps (see figure \ref{Regimes}). At first glance, to resolve this problem one might think to increase the attraction between BMU and its neighbors. However, this approach will not work for symmetry reasons. Inspecting the tangled maps from experiment 1, in figure 2, reveals that they all share a common attribute of having non-uniform errors values across their units.

Therefore, we propose a feedback mechanism that only allows for maps with a uniform level of quality to remain stationary. We achieve this by coupling the neighborhood attraction of a BMU to its quantization and alfa errors. By doing so, defective segments of the map increase their attraction, resulting in the flow of units towards them to prevent stationarity. Defining what constitute a defective segment requires meaningful aggregation of different errors (quantization, alfa). We refer to this SOM variant as Feedback NNSOM (FNNSOM).

This feedback mechanism can be implemented by a function which increases the neighborhood attraction of units which suffer either from high quantization or alpha error. This requires that the  learning rate is a strictly increasing function of alfa and quantization error. Defining such a function requires aggregating map alfa and quantization errors. However, this is problematic since alfa error is bounded between 0 and 1, while quantization error depends on the data and is unbounded. Therefore, to compare the two errors rates, quantization error needs to be mapped to the same interval. The requirements described above are satisfied by the following equations:

\begin{subequations}

\begin{gather}
f_j\big(i \big)=1 - \exp \left( - \frac {c_q  \bar q_{b_i}(i) } { \bar q_j(i)} \right)  \label{qfeedback} \\
F_j\big(i \big) =     \bar{\alpha}_{b_i}(i)+f_j(i)- \bar{\alpha}_{b_i}f_j(i) \label{feedbackfnc}
\end{gather}
\begin{align} \label{feq}	   
r_{{j}} \big(i \big) &=
\begin{cases}
\sigma & \text{ if $j$ is $b_i$}  \\
F_j \big(i \big) & \text{ if j is a neighbor of $b_i$}   \\
0 & \text{ otherwise}  \\
\end{cases}
\end{align}
\end{subequations}
\begin{gather}
\sigma \text{ is a user defined constant for the BMU learning rate }\notag\\
c_q  \text{ is a user-configured parameter that tunes contribution}\notag\\ \text{of quatization error at the unit's learning rate} \notag
\end{gather}
Equation \eqref{qfeedback} is a strictly increasing function of the quantization error of the BMU bounded between 0 and 1. We normalize the BMU quantization error by expressing it as a fraction of the quantization error of the unit $j$. The sensitivity of the function to the normalized quantization error is controlled by the user defined parameter $c_q$. Equation \eqref{feedbackfnc} is a strictly increasing function of both alfa and quantization errors \footnote{please note that the quantization error dependence is through $f_j(i)$} of the BMU. Additionally, equation \eqref{feedbackfnc} is bounded between 0 and 1 so that it can act as a learning rate, and either error can independently result in a maximum value of 1. Equation \eqref{feq} states 3 cases for the learning rate: a user-defined constant $\sigma$ for the BMU, feedback determined for its neighbors, and 0 otherwise. For this work, $\sigma$ is set to $0.01$, however, other parameterizations are not ruled out. 
	\begin{figure*}
		\centering
		 \includegraphics[width=\textwidth,scale=0.4]{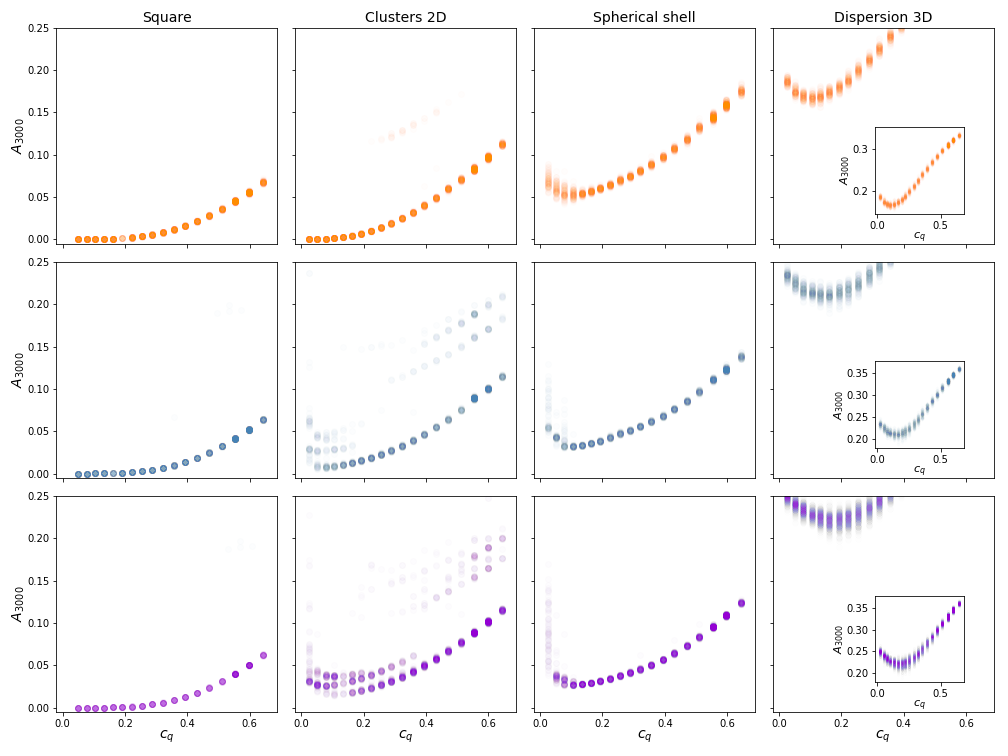}
		\caption{illustrates the map alfa error at iteration 3000 of FNNSOM. We consider four datasets (Square, Clusters 2D, Spherical shell and Dispersion 3D) with random initial conditions, over a range of 19 $c_q$ values (see eq. \eqref{qfeedback}), for increasing map sizes from top to bottom (100, 400, 900). Each combination is run 100 times. Note that only dataset with inconsistent map alfa error is Clusters 2D. }\label{FNNSOM_1}
	\end{figure*}

The efficacy of the suggested algorithm (FNNSOM) is assessed empirically over synthetic datasets. We picked four diverse datasets, see table \ref{datasets}, to investigate whether FNNSOM suffers from robust topological defects. We are also interested in how the algorithm performs with increasing system size. We consider three individual maps sizes $(100, 400, $ $ 900)$ where the $c_q$ parameter is sampled geometrically at $19$ points between $0$ and $1$.
	
Each combination of the parameters above is simulated $100$ times, for $3000$ iterations, and with random initial conditions (RIC).
Observing the values of $A_{3000}$ reveals that the performance of FNNSOM strongly depends on the dataset, as illustrated in figure \ref{FNNSOM_1} . Specifically:
\begin{enumerate}
	\item
	For all datasets, except Clusters 2D, the values of $A_{3000}$ are all clustered around the same level, with very few exceptions residing in the cloud of tangled maps.
	Manual inspection of the trained maps reveals that they all have a correct global topology.
	The lowest value of $A_{3000}$ is encountered for $c_q$ approximately around $0.15$, as indicated in figure \ref{FNNSOM_1}.
	We therefore conclude that for the considered datasets, and for a band of $c_q$ values, the FNNSOM is highly resistant to tangling, and a significant improvement over the NNSOM
		\item
	The Clusters 2D dataset is an exception to the previous point. Specifically, we observed that $A_{3000}$ values are not tightly clustered, instead they are dispersed across multiple levels. On manual inspection, we found that higher $A_{3000}$ values indeed correspond to the topologically deformed maps. This behavior exacerbated with increasing system size, as seen in the figure \ref{FNNSOM_1} where more points can be found at higher $A_{3000}$ as system size increases.
	\item
	Increasing map size has a detrimental effect in all data sets but the Square and Spherical shell datasets. As figure \ref{FNNSOM_1} shows that for Square and Spherical shell, increasing system size decreases $A_{3000}$ for all $c_q$ values. The opposite is seen for the remaining datasets.	
\end{enumerate}
	\begin{table} 
		\begin{tabular}{ll}
			Square          & $s_i$ sampled at random from $[0,1]^2$                                                                                                                                     \\
			Clusters 2D     & \begin{tabular}[c]{@{}l@{}}$s_i$ sampled at random from five identical\\ 2D normal  multivariate distributions at \\ $(0,0),~(0,5),~(5,0),~(5,5),~(2.5, 2.5)$\end{tabular} \\
			Spherical Shell & \begin{tabular}[c]{@{}l@{}}$s_i$ sampled uniformly at random on the\\ surface of a  sphere\end{tabular}                                                                    \\
			Dispersion      & \begin{tabular}[c]{@{}l@{}}$s_i$ sampled from a 3D normal, multinomial\\ distribution\end{tabular}                                                                       
		\end{tabular}	
		\caption{Description of the datasets used to assess FNNSOM in sec. \ref{fnnsomexp} }\label{datasets}	 
	\end{table}	
In conclusion, the suggested feedback parameterization is very effective in treating tangles in all considered datasets except Clusters 2D. The performance of FNNSOM was observed to be optimal over a range of values of $c_q$, which from a practical standpoint means that the algorithm does not require precise tuning. Additionally, the effect of increasing system size on the eventual $A_{3000}$ is either detrimental or positive - depending on the dataset type. 

	The inability of FNNSOM to consistently resolve Clusters 2D can be intuitively understood with a simple thought experiment. Consider 2 clusters of data, separated by a large gap in which no sample ever arrives. The map units will split between the two data clusters, and thus some units will be inevitably placed over the gap.	These units residing over the gap will not be receiving samples, and will therefore never attract their neighbors. Consequently, these units are unable to relay any attraction across the two unit populations residing in each data cluster.
	Therefore, these two unit populations are effectively acting independently and thus unable to collaborate to resolve global topological deformation.

Thus, the FNNSOM seem to be particularly well suited for datasets over 2D manifolds. 
In fact, the performance of FNNSOM over such datasets scales along with system size.

\section{Applicability Analysis}\label{:scenario}
 
 To establish the practicality of the algorithm, we investigate whether FNNSOM can reach a  stable stationarity state for some $c_q$ values. Additionally, we test for scalability using a dataset from an industrial application.

\begin{figure}
 	\includegraphics[scale=0.55]{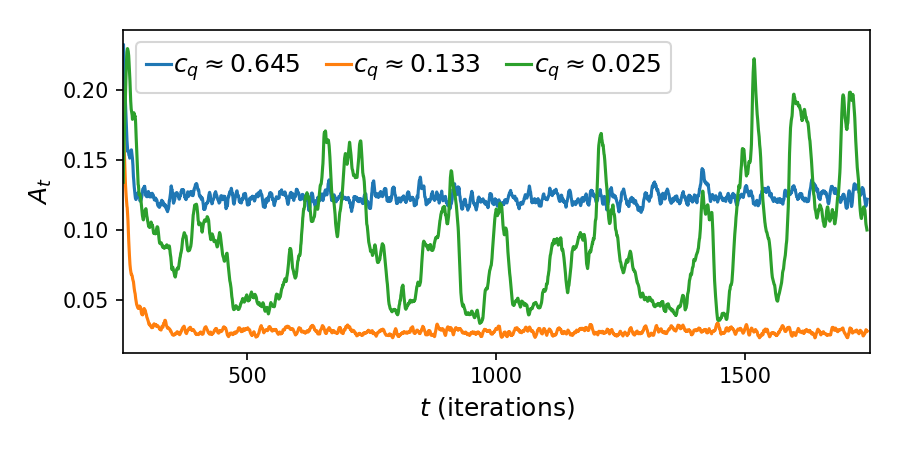}
 	\caption{depicts behavior of FNNSOM over a spherical shell, for 900 map size and for three different $c_q$ values.
 		We illustrate the evolution of the map alfa error throughout the training process for the 3 $c_q$ values as listed in the legend. 
 		For the highest $c_q$, the map reaches stationarity at high error, whereas lowering $c_q$ to an intermediate range (e.g., 0.133) results in a significantly lower error value.
 		For the lowest $c_q$, the variability of error increases drastically, demonstrating that the map never reaches a fixed steady state.
 	 	}\label{FNNSOM_2}
 	
 \end{figure}

 \textbf{Stationarity:} We investigate the influence of $c_q$ over the behavior of the map by observing $A_{t}$ for varying $c_q$ values and for $3000$ iterations, as depicted in figure \ref{FNNSOM_2}. We apply the algorithm over the Spherical shell dataset, with the map size fixed at $900$. The $c_q$ parameter is sampled in  $\{0.025, 0.133,$ $ 0.645\}$ which correspond to the local extrema of $A_{3000}$, as identified from figure \ref{FNNSOM_1} (bottom row, third column).

The results reveal that $c_q$ affects both map stability and topological preservation.
 As depicted in figure \ref{FNNSOM_2}, for $c_q = 0.025$, $A_t$ varies wildly in time - indicating that the map is unable to reach a stable state.
 This persistent variability of  $A_t$ for low $c_q$ values is consistent with the widely dispersed $A_{3000}$ for $c_q=0.025$ in the previous experiments as seen in figure \ref{FNNSOM_1} (third column). In contrast the two higher $c_q$ values reach a near constant level of $A_t$ within the first few hundred iterations. There is however significant difference between the two levels, with $c_q=0.133$ having approximately one fourth of the $A_t$ at $c_q=0.645$. The higher $A_t$ at $c_q=0.645$ results from a noisy map (shown in figure \ref{Regimes}, C2). These observations imply that the identified range of near optimal $c_q$ (shown in figure \ref{FNNSOM_2}) tunes the map between instability and topological deformation.

 \textbf{Scalability:} We test the applicability and scalability of the algorithm with the commonly  used \cite{maxplanckcite,isgro2005clustering} dataset of Max Planck's head \footnote{\url{http://visionair.ge.imati.cnr.it/ontologies/shapes/view.jsp?id=77-Max-Planck_bust}}. The dataset is a point could of $200k$ points. As a point cloud dataset, it represents a 3D object as number of points residing over a 2D manifold, and FNNSOM is therefore well suited for this problem. In addition, we can visually inspect the quality of the resulting map by plotting it as a mesh. Finally, the large number of data points allows us to test scalability.

\begin{figure*}
 	\centering
 	\includegraphics[scale=1]{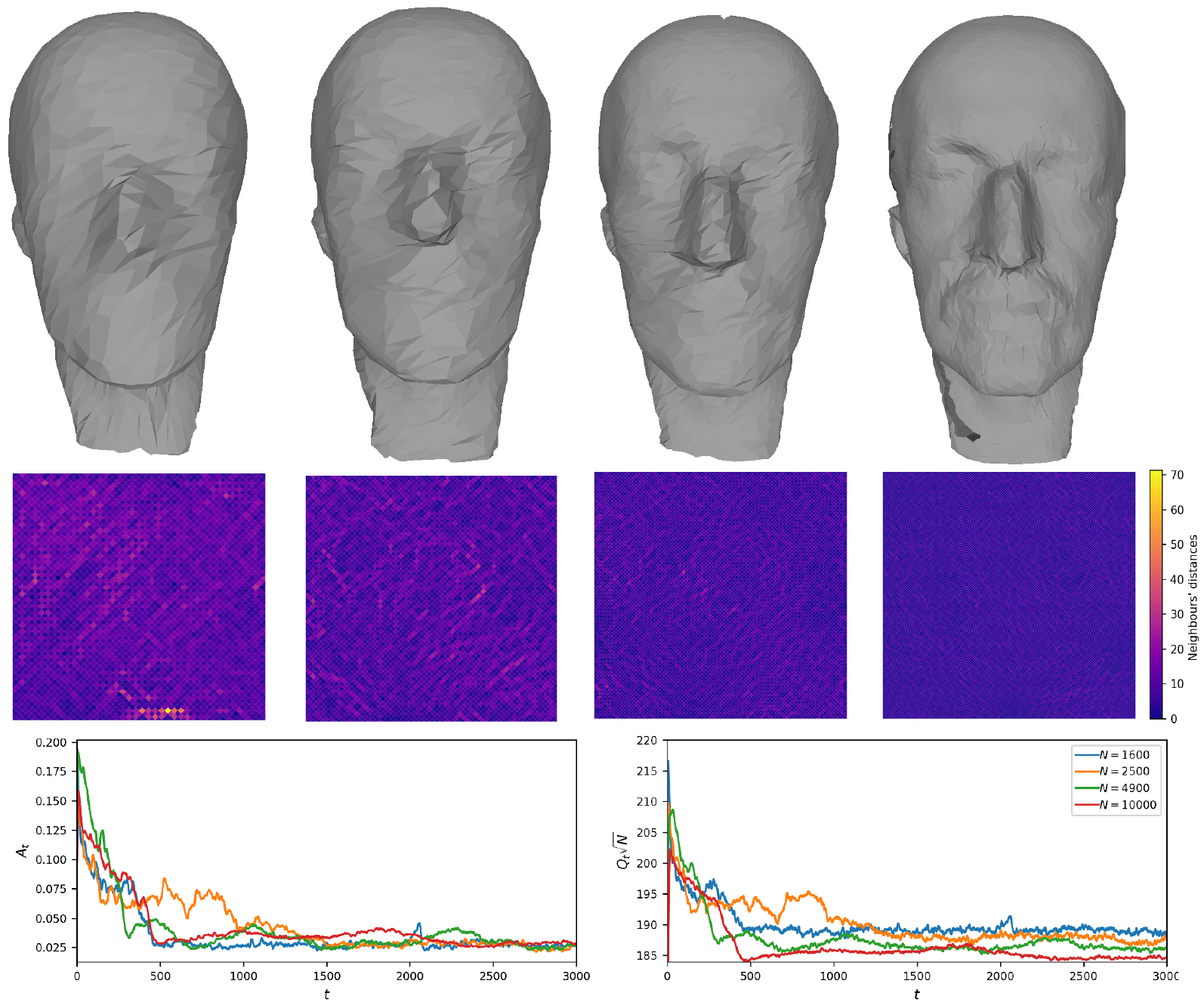}
 	\caption{depicts the scalability of the FNNSOM algorithm by training maps of increasing size over a standard dataset of 200k data-points (point-cloud of Max Planck's head) for $c_q = 0.133$
 		\textbf{Top row:} We visualize the maps at iteration 3000 using 3d polygons. Vertices are positioned at the map unit locations, and triangles are shapes between neighboring units. 
 		Sizes of the map increase from left to right ($1.6k, 2.5k 4.9k, 10k$). 
 		Note that the resultant resolution increases along with the map size.
 		\textbf{Second row:} We show u-matrices for the corresponding maps sizes. 	
 		Distances between neighboring units decrease uniformly with increasing map sizes and no topological defects appear.  		
 		\textbf{Bottom row:}  We visualize the average alpha ($A_t$) and quantization ($Q_t$) errors for each iteration of the maps above. $Q_t$ is normalized by multiplying with square root of the map size ($\sqrt{N}$). All maps follow similar trajectory regardless of their sizes. This shows that the number of required samples scales linearly with the maps size.}\label{MAXPLANCK}
 \end{figure*}

We use maps of sizes ($1.6k, 2.5k 4.9k, 10k$), for $3000$ iterations, and for $c_q = 0.133$. In figure \ref{MAXPLANCK}, top row, we depict the meshes of the trained maps. We observe no tangles in the meshes, as it is corroborated by the respective U-Matrices in the second row.

The scalability is observed by increasing resolution of the meshes as well as increasing uniformity of the U-Matrices. Additionally, in the bottom row of the figure \ref{MAXPLANCK}, we observe that the map converges to the same eventual errors, at approximately the same rate - regardless of size. This implies a linear relationship between the number of training samples needed and the size of the map.

\section{Discussion}\label{:discussion}

The challenge in locally interacting units is uncovering the global topology of the data, as confirmed by the experiment in section \ref{CLRSec}. Our results verify that locally interacting self organizing maps (SOM)s suffer from tangles: robust, global topological defects, where the map units are characterized by inhomogeneous error values.

We have demonstrated that imposing a homogeneous error value throughout the map is possible using localized feedbacks. Specifically, our results show that such a feedback allows the map to resolve tangles in three, diverse synthetic, datasets while it is unable to consistently treat topological deformation for sparse datasets. Additionally, we establish the applicability of the algorithm by deriving a range for the hyperparameter used in the feedback function, and by demonstrating its scalability on an industrial application.

The suggested algorithm allows locally interacting SOMs to discover global topology. This approach drastically reduces computational complexity by minimizing neighborhood size. Additionally, small neighborhoods are a first step towards a truly decentralized implementation of SOM, where the algorithm is executed simultaneously over multiple machines with embarrassingly parallel simplicity. Finally, the algorithm decouples the neighborhood attraction from time (number of samples processed).

Future work could focus on a more formal description of the abrupt loss of stability of the map for low $c_q$ values, as well as on resolving tangles in the case of sparse data. Additionally, modifications that would enable an embarrassingly parallel SOM could be investigated, most likely focusing on finding an alternative for the best matching unit search. Finally, different initialization strategies suited to locally interacting SOMs could be investigated.

\section*{Acknowledgment}

The research published here was conducted at the Future Resilient Systems at the Singapore-ETH Centre (SEC). The SEC was established as a collaboration between ETH Zurich and National Research Foundation (NRF) Singapore (FI 370074011) under the auspices of the NRF's Campus for Research Excellence and Technological Enterprise (CREATE) programme. 

\bibliographystyle{elsarticle-harv}
\bibliography{FNNSOM}

\begin{thebibliography}{12}
\expandafter\ifx\csname natexlab\endcsname\relax\def\natexlab#1{#1}\fi
\expandafter\ifx\csname url\endcsname\relax
  \def\url#1{\texttt{#1}}\fi
\expandafter\ifx\csname urlprefix\endcsname\relax\def\urlprefix{URL }\fi

\bibitem[{Cottrell et~al.(1998)Cottrell, Fort, and
  Pag{\`e}s}]{LRMapConvergence}
Cottrell, M., Fort, J.-C., Pag{\`e}s, G., 1998. Theoretical aspects of the som
  algorithm. Neurocomputing 21~(1-3), 119--138.

\bibitem[{Isgro et~al.(2005)Isgro, Odone, Saleem, and
  Schall}]{isgro2005clustering}
Isgro, F., Odone, F., Saleem, W., Schall, O., 2005. Clustering for surface
  reconstruction. In: 1st International Workshop on Semantic Virtual
  Environments. pp. 156--162.

\bibitem[{Ivrissimtzis et~al.(2003)Ivrissimtzis, Jeong, and
  Seidel}]{pointcloud_3}
Ivrissimtzis, I., Jeong, W.-K., Seidel, H.-P., 2003. Using growing cell
  structures for surface reconstruction. In: Shape Modeling International,
  2003. IEEE, pp. 78--86.

\bibitem[{Keith-Magee et~al.(1999)Keith-Magee, Venkatesh, and
  Takatsuka}]{keith1999_empirical}
Keith-Magee, R., Venkatesh, S., Takatsuka, M., 1999. An empirical study of
  neighbourhood decay in kohonen's self organizing map. In: IJCNN 1999:
  Proceedings of the International Joint Conference on Neural Networks. IEEE,
  pp. 1953--1958.

\bibitem[{Kohonen(2013)}]{kohonen2013_essentials}
Kohonen, T., 2013. Essentials of the self-organizing map. Neural networks 37,
  52--65.

\bibitem[{Kohonen et~al.(2000)Kohonen, Kaski, Lagus, Salojarvi, Honkela,
  Paatero, and Saarela}]{massivemap}
Kohonen, T., Kaski, S., Lagus, K., Salojarvi, J., Honkela, J., Paatero, V.,
  Saarela, A., May 2000. Self organization of a massive document collection.
  IEEE Transactions on Neural Networks 11~(3), 574--585.

\bibitem[{Kohonen et~al.(2009)Kohonen, Nieminen, and Honkela}]{KohonenQError}
Kohonen, T., Nieminen, I.~T., Honkela, T., 2009. On the quantization error in
  som vs. vq: A critical and systematic study. In: Proceedings of the 7th
  International Workshop on Advances in Self-Organizing Maps. WSOM '09.
  Springer-Verlag, Berlin, Heidelberg, pp. 133--144.

\bibitem[{Kohonen et~al.(2001)Kohonen, Schroeder, and Huang}]{kohonen_book}
Kohonen, T., Schroeder, M.~R., Huang, T.~S. (Eds.), 2001. Self-Organizing Maps,
  3rd Edition. Springer-Verlag, Berlin, Heidelberg.

\bibitem[{Li et~al.(2018)Li, Chen, and Hee~Lee}]{pointcloud_1}
Li, J., Chen, B.~M., Hee~Lee, G., 2018. So-net: Self-organizing network for
  point cloud analysis. In: Proceedings of the IEEE conference on computer
  vision and pattern recognition. pp. 9397--9406.

\bibitem[{Pauly et~al.(2006)Pauly, Kobbelt, and Gross}]{maxplanckcite}
Pauly, M., Kobbelt, L.~P., Gross, M., Apr. 2006. Point-based multiscale surface
  representation. ACM Trans. Graph. 25~(2), 177--193.

\bibitem[{Rego et~al.(2010)Rego, Ara{\'u}jo, and de~Lima~Neto}]{pointcloud_2}
Rego, R. L. M.~E., Ara{\'u}jo, A. F.~R., de~Lima~Neto, F.~B., 2010. Growing
  self-reconstruction maps. IEEE transactions on neural networks 21~(2),
  211--223.

\bibitem[{Uriarte and Mart{\'\i}n(2005)}]{alfaError}
Uriarte, E.~A., Mart{\'\i}n, F.~D., 2005. Topology preservation in som.
  International journal of applied mathematics and computer sciences 1~(1),
  19--22.

\end{thebibliography}

\end{document}